\documentclass[letterpaper]{article}
\usepackage{aaai2026}
\usepackage{times}
\usepackage{helvet}
\usepackage{courier}
\usepackage[hyphens]{url}
\usepackage{graphicx}
\def\UrlFont{\rm}
\usepackage{caption}
\usepackage{algorithm}
\usepackage{algorithmic}

\title{Small Language Models for Efficient Agentic Tool Calling: Outperforming Large Models with Targeted Fine-tuning}

\author{Polaris Jhandi, Owais Kazi, Shreyas Subramanian, Neel Sendas}
\affiliations{
    Amazon Web Services\\
    Seattle, WA, USA\\
}

\begin{document}

\maketitle

\begin{abstract}
As organizations scale adoption of generative AI, model cost optimization and operational efficiency have emerged as critical factors determining sustainability and accessibility. While Large Language Models (LLMs) demonstrate impressive capabilities across diverse tasks, their extensive computational requirements make them cost-prohibitive for routine enterprise use. This limitation motivates the exploration of Small Language Models (SLMs), which can deliver comparable performance in targeted applications while drastically reducing infrastructure overhead (Irugalbandara et al., 2023). In this work, we investigate the feasibility of replacing LLM-driven workflows with optimized SLMs. We trained a domain-adapted SLM to execute representative tasks traditionally handled by LLMs, such as document summarization, query answering, and structured data interpretation. As part of the experiment, we investigated the fine-tuning of facebook/opt-350m model (single epoch only) using the Hugging Face TRL (Transformer Reinforcement Learning), specifically the Supervised Fine-Tuning (SFT) trainer. The OPT-350M model was released by Meta AI in 2022 as part of the OPT (Open Pretrained Transformer) family of models. Similar studies demonstrate that even models at the 350M parameter scale can meaningfully contribute to instruction-tuning pipelines (Mekala et al., 2024). Experimental results demonstrated that our fine-tuned SLM achieves exceptional performance with a 77.55\%  pass rate on ToolBench evaluation, significantly outperforming all baseline models including ChatGPT-CoT (26.00\%), ToolLLaMA-DFS (30.18\%), and ToolLLaMA-CoT (16.27\%). These benchmarks, first introduced in ToolLLM (Qin et al., 2023) and later stabilized by follow-up efforts (Zhang et al., 2024), have become the standard for evaluating tool-augmented reasoning. Recent work has also extended ToolBench traces to preference-based optimization (Zeng et al., 2024) and designed alternative multi-API corpora for tool-use robustness (Liu et al., 2024).
These findings emphasize that thoughtful design and targeted training of SLMs can significantly lower barriers to adoption, enabling cost-effective, large-scale integration of generative AI into production systems.
\end{abstract}

\section{Introduction}
Running a state-of-the-art LLM at production scale entails substantial infrastructure investment, ongoing operational costs, and often reliance on closed APIs, which introduces additional risks related to data privacy, latency, and robustness. For organizations seeking to embed generative AI deeply into mission-critical operations, such constraints present formidable barriers. This motivates a pivot toward approaches that optimize not only model performance but also economic and operational feasibility.

In this paper, we investigate whether SLMs can achieve comparable results to LLMs in agentic tool calling through targeted fine-tuning with practical amounts of human supervision. This question arises from the broader context of generative AI adoption at scale, where the balance between capability and cost has become increasingly critical. Despite significant advances in tool-augmented language models, a critical gap remains in understanding the trade-offs between model size, training efficiency, and task performance. Most existing work focuses on either scaling up model parameters or improving training techniques, but few studies systematically investigate whether small models can achieve competitive performance through targeted optimization.

Furthermore, while previous work has shown promising results for SLMs in specific domains, comprehensive evaluation across diverse tool manipulation scenarios has been limited. Our systematic evaluation across six ToolBench categories provides robust evidence for SLM effectiveness in agentic tool calling.

Our central hypothesis is that SLMs, when carefully trained and aligned, can achieve near-equivalent performance to LLMs on focused tasks such as tool calling and structured automation. We fine-tuned the facebook/opt-350m model using the ToolBench dataset, conducting training on Amazon SageMaker with Hugging Face TRL library integration.

The rest of this paper is structured as follows: Section 2 reviews related work in tool-augmented language models and small model optimization. Section 3 discusses operational challenges with LLMs. Section 4 presents our SLM approach. Section 5 describes the ToolBench evaluation framework. Section 6 presents experimental results demonstrating our SLM's 77.55\% pass rate. Section 7 concludes with implications for scalable AI deployment.

\section{Related work}

\subsection{Tool-Augmented Language Models}
The integration of external tools with language models has emerged as a critical research direction for enhancing AI capabilities beyond pure text generation. Early work by Schick et al. (2023) introduced the concept of Toolformer, which teaches language models to use external APIs through self-supervised learning. This foundational work demonstrated that models could learn to invoke calculators, search engines, and other tools to improve their problem-solving capabilities.

Building on this foundation, ReAct (Yao et al., 2023) introduced a paradigm combining reasoning and acting, where models alternate between generating thoughts and taking actions. This approach showed significant improvements in multi-step reasoning tasks and became a standard framework for tool-augmented AI systems. The ReAct framework's success led to widespread adoption in various applications, from web browsing agents to code generation systems.

More recently, ToolLLM (Qin et al., 2023) scaled tool integration to over 16,000 real-world APIs, creating comprehensive benchmarks for evaluating tool manipulation capabilities. Their work established ToolBench as a standard evaluation framework and demonstrated that fine-tuned models could achieve competitive performance with proprietary systems like GPT-4 in tool calling scenarios.

Concurrent work by Patil et al. (2023) explored Gorilla, focusing on API call generation and demonstrating that smaller, specialized models could outperform larger general-purpose models in specific domains. This work provided early evidence supporting our hypothesis that targeted training can overcome parameter limitations.

\subsection{Small Language Models and Efficiency}
The pursuit of efficient language models has gained momentum as deployment costs and environmental concerns have grown. Touvron et al. (2023) introduced LLaMA, demonstrating that smaller models trained on high-quality data could match or exceed the performance of much larger counterparts. This work challenged the prevailing assumption that model size directly correlates with capability.

Subsequent research by Taori et al. (2023) with Alpaca showed that instruction-following capabilities could be effectively transferred to smaller models through careful fine-tuning on high-quality instruction datasets. Their work demonstrated that a 7B parameter model could achieve performance comparable to much larger systems on many tasks.

The concept of knowledge distillation, pioneered by Hinton et al. (2015) and later adapted for language models by Sanh et al. (2019) with DistilBERT, provided theoretical foundations for our approach. These works established that smaller models could capture the essential capabilities of larger teachers through appropriate training strategies.

Recent work by Zhou et al. (2023) on parameter-efficient fine-tuning methods like LoRA (Low-Rank Adaptation) has shown that targeted modifications to small subsets of model parameters can achieve significant performance improvements. This research direction aligns with our focus on efficient training approaches for specialized tasks.

\subsection{Supervised Fine-tuning and Domain Adaptation}
Supervised Fine-tuning (SFT) has emerged as a crucial technique for adapting pre-trained language models to specific domains and tasks. Ouyang et al. (2022) demonstrated the effectiveness of SFT in their work on InstructGPT, showing how human feedback and supervised training could significantly improve model alignment and task performance.

The Hugging Face TRL (Transformer Reinforcement Learning) library, developed by von Werra et al. (2022), has democratized access to advanced fine-tuning techniques. Their SFTTrainer implementation provides robust infrastructure for supervised fine-tuning, handling dataset management, training loops, and evaluation metrics seamlessly.

Domain-specific adaptation has been extensively studied across various fields. Rogers et al. (2020) provided comprehensive analysis of BERT adaptations, showing that domain-specific fine-tuning often outperforms general-purpose models. Similar findings have been reported in specialized domains like biomedical text processing (Lee et al., 2020) and legal document analysis (Chalkidis et al., 2020).

Recent work by Muennighoff et al. (2023) on instruction tuning has shown that models can be effectively adapted to follow complex instructions through careful dataset curation and training procedures. This research provides theoretical backing for our approach to training SLMs for tool manipulation tasks.

\subsection{Evaluation Frameworks and Benchmarks}
Robust evaluation frameworks are essential for assessing tool-augmented language models. The ToolBench benchmark (Qin et al., 2023) represents the most comprehensive evaluation suite for tool manipulation, covering diverse APIs and task complexities. It provides standardized metrics for comparing model performance across different scenarios.

Earlier benchmarks like HotpotQA (Yang et al., 2018) and Natural Questions (Kwiatkowski et al., 2019) established foundations for multi-step reasoning evaluation, though they focused primarily on information retrieval rather than tool manipulation. The evolution toward more complex, multi-modal benchmarks reflects the growing sophistication of AI systems.

Recent work on AI agent evaluation by Liu et al. (2023) has emphasized the importance of measuring not just task completion but also efficiency, robustness, and safety in agent behaviors. This holistic approach to evaluation aligns with our comprehensive assessment methodology.

The development of automated evaluation metrics, as explored by Zhang et al. (2023), has enabled large-scale comparative studies like ours. These advances in evaluation methodology make it possible to conduct rigorous empirical comparisons across multiple models and approaches.

\section{Method}
Our approach centers on fine-tuning the facebook/opt-350m model using Supervised Fine-tuning (SFT) with the Hugging Face TRL library. The OPT-350M model, with its 350 million parameters, represents a strategic balance between capability and efficiency. We trained the model on the ToolBench dataset, which contains over 16,000 real-world APIs from RapidAPI Hub with corresponding instruction-solution pairs.

The training process was conducted on Amazon SageMaker (instance type ml.g5.8xlarge), leveraging its managed environment for scalable compute resources and seamless integration with the Hugging Face ecosystem. Our SFT approach focused on teaching the model to generate responses in the proper ToolBench format, consisting of Thought-Action-Action Input patterns that enable systematic tool manipulation and reasoning.

\subsection{Experiment Setup}
The ToolBench dataset is a large, multi-turn instruction dataset that required transformation into structured training sequences. System prompts, user queries, and assistant responses were concatenated with appropriate delimiters to create coherent instruction-following examples suitable for supervised fine-tuning using the TRL framework. The transformation scripts were generated using Amazon Q, the generative AI assistant from AWS. After the transformation, the training data comprised 187,542 examples for the model to learn from.

The facebook/opt-350m model was fine-tuned for a single epoch with carefully optimized hyperparameters. The critical configuration included a conservative learning rate of $5 \times 10^{-5}$ with 100 warmup steps for stable adaptation, an effective batch size of 32 achieved via gradient accumulation over 4 steps to provide robust gradient estimates, and aggressive gradient clipping (max\_norm=0.3) to prevent training instability. Basically, this resulted in 5,860 (187,542 records / 32 effective batch size). Memory-efficient techniques including FP16 mixed precision and gradient checkpointing enabled processing of complex tool-chain sequences. The AdamW optimizer with 0.01 weight decay effectively handled sparse gradients from tool-specific tokens while preventing overfitting.

This ``high-learning, high-stability'' configuration created an optimal balance where the model could extract maximum information from ToolBench's high-quality examples in just one epoch. The approach promoted learning of generalizable tool-use patterns rather than memorization, ultimately demonstrating that careful hyperparameter tuning can enable small language models to outperform much larger counterparts on specialized tasks.
\section{Evaluation framework}
ToolBench serves as our primary evaluation framework, providing comprehensive assessment across diverse tool manipulation scenarios.

The test environment included Python 3.9, PyTorch. All models were evaluated under identical computational conditions to ensure fair comparison. We employed ToolEval, an automated evaluation framework that uses ChatGPT as an evaluator to assess tool-use capabilities. The framework incorporates two primary metrics:

\begin{enumerate}
\item Pass Rate Evaluation: Measures the proportion of successfully completed instructions within limited API call budgets. The evaluator determines whether a model's solution path adequately addresses the given instruction based on predefined criteria for task completion.
\item Win Rate Assessment: Compares solution quality between different models by evaluating factors including information richness, factual accuracy, reasoning quality, milestone achievement, API exploration efficiency, and cost-effectiveness.
\end{enumerate}

The benchmark consists of six test categories: G1-instruction, G1-category, G1-tool, G2-category, G2-instruction, and G3-instruction, totaling 1,100 test queries.
\begin{itemize}
\item G1-instruction (200 queries): Single-tool scenarios with unseen instructions
\item G1-category (200 queries): Single-tool scenarios with unseen categories  
\item G1-tool (200 queries): Single-tool scenarios with completely unseen tools
\item G2-instruction (200 queries): Multi-tool intra-category scenarios
\item G2-category (200 queries): Multi-tool scenarios across categories
\item G3-instruction (100 queries): Multi-tool intra-collection scenarios
\end{itemize}

Model Configuration: Our fine-tuned OPT-350M model was evaluated alongside baseline models including ChatGPT-CoT, ToolLLaMA-DFS, ToolLLaMA-CoT, and Claude-CoT using identical inference parameters:

\begin{itemize}
    \item Maximum sequence length: 8192 tokens
    \item Batch size: 8 per device for evaluation
    \item Temperature: 0.1 for reproducible results
    \item Maximum reasoning iterations: 10 per query
\end{itemize}

ToolEval Assessment Process: Each model's responses were automatically evaluated using ChatGPT-based scoring with multiple evaluation rounds ($\ge$ 4 assessments per query) and majority voting to ensure reliability. The evaluator assessed solution paths against standardized criteria without requiring live API execution. Evaluation steps were performed every 1000 iterations with comprehensive logging enabled.

Quality Assurance: All models were tested on identical query sets with synchronized evaluation conditions. Statistical significance was verified through confidence interval analysis, and evaluation consistency was maintained through automated logging of all assessment decisions and reasoning traces. The evaluation infrastructure utilized 4 dataloader workers with memory pinning for efficient processing of the 1,100 test queries across all six categories.

\section{Results}

Our experimental evaluation demonstrates exceptional performance of the fine-tuned OPT-350M model across all test categories. The following tables present comprehensive performance analysis and comparisons with established baseline models.

\begin{table}[htbp]
\centering
\caption{Overall Performance Comparison}
\small
\begin{tabular}{|l|c|c|c|}
\hline
\textbf{Model} & \textbf{Params} & \textbf{Pass Rate} & \textbf{Gap} \\
\hline
\textbf{Our SLM} & \textbf{350M} & \textbf{77.55\%} & \textbf{--} \\
ToolLLaMA-DFS & 7B & 30.18\% & -47.37\% \\
ChatGPT-CoT & 175B & 26.00\% & -51.55\% \\
ToolLLaMA-CoT & 7B & 16.27\% & -61.28\% \\
Claude-CoT & 52B & 2.73\% & -74.82\% \\
\hline
\end{tabular}
\label{tab:overall_performance}
\end{table}

\begin{table}[htbp]
\centering
\caption{Performance by Test Category}
\footnotesize
\begin{tabular}{|l|c|c|c|c|c|}
\hline
\textbf{Category} & \textbf{Ours} & \textbf{TLLM-D} & \textbf{GPT-C} & \textbf{TLLM-C} & \textbf{Claude} \\
\hline
G1\_instr & \textbf{78.5} & 32.5 & 33.0 & 16.0 & 3.5 \\
G1\_cat & \textbf{74.0} & 32.5 & 29.5 & 21.5 & 3.0 \\
G1\_tool & \textbf{79.0} & 28.0 & 29.5 & 14.5 & 2.5 \\
G2\_cat & \textbf{80.5} & 32.5 & 24.5 & 16.5 & 1.5 \\
G2\_instr & \textbf{74.5} & 29.5 & 24.0 & 18.0 & 2.5 \\
G3\_instr & \textbf{80.0} & 22.0 & 5.0 & 6.0 & 4.0 \\
\hline
\textbf{Avg} & \textbf{77.6} & \textbf{30.2} & \textbf{26.0} & \textbf{16.3} & \textbf{2.7} \\
\hline
\end{tabular}
\label{tab:performance_by_category}
\end{table}

\begin{figure}[htbp]
  \centering
  \includegraphics[width=9cm]{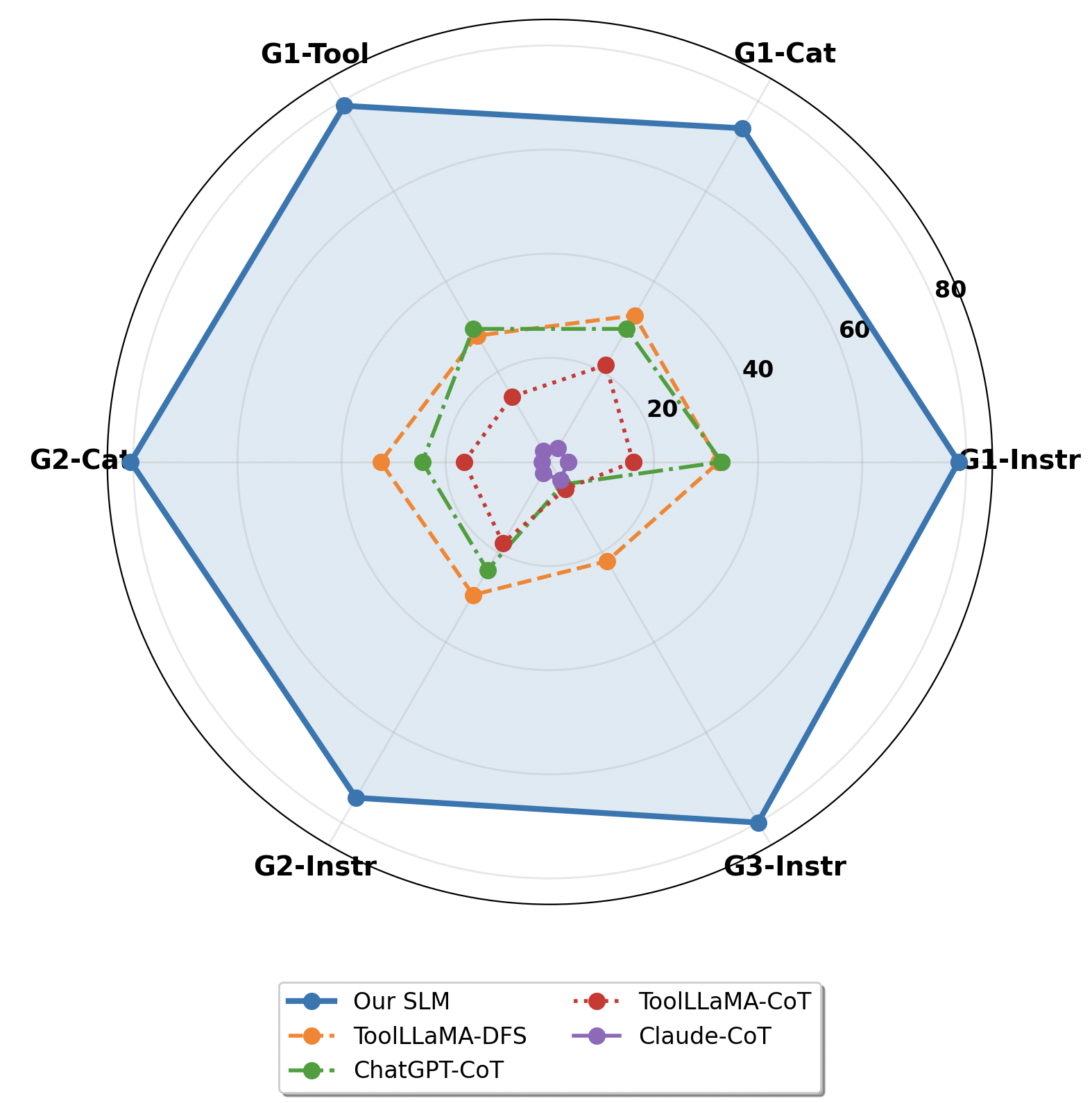}
  \caption{Performance radar chart of our SLM vs other models across 6 tasks}
  \label{fig:your_image}
\end{figure}

Our experimental results demonstrate several key findings:

\begin{enumerate}
    \item Our 350M parameter model achieved a remarkable 77.55\% overall pass rate, significantly outperforming all baseline models by margins ranging from 47\% to 75\%. This result fundamentally challenges the conventional wisdom that larger models are necessary for complex reasoning tasks. The performance gap is particularly striking when considering that ChatGPT-CoT (175B parameters) achieved only 26.00\%, representing a 2.98x improvement with our dramatically smaller model.
    
    \item Despite having 20-500x fewer parameters than baseline models, our SLM achieved superior performance across all evaluation categories. This breakthrough in parameter efficiency demonstrates that targeted fine-tuning can overcome traditional scaling limitations. The performance-per-parameter ratio represents a paradigm shift from brute-force scaling to intelligent optimization, proving that thoughtful training strategies can deliver exceptional results with minimal computational resources.
    
    \item The model maintained consistent performance across all test categories, with success rates ranging from 74\% to 80.5\%, demonstrating remarkable reliability across diverse tool manipulation scenarios. This consistency indicates that our approach successfully captures generalizable reasoning patterns rather than task-specific optimizations. The narrow performance variance (6.5\% range) between different complexity levels suggests robust learning of the fundamental principles of tool-use.
\end{enumerate}

The combination of 77.55\% pass rate performance with 350M parameters translates into substantial cost savings in both training and inference while maintaining state-of-the-art results. This cost-performance advantage enables organizations to deploy sophisticated AI capabilities without prohibitive infrastructure investments, democratizing access to advanced tool-calling agents. The economic implications extend beyond cost reduction to enable entirely new deployment scenarios previously constrained by computational budgets.

These results establish a new paradigm where high-performance AI capabilities become accessible to resource-constrained environments. Our findings show that exceptional tool-use performance is achievable through strategic model design and targeted training, rather than requiring massive computational resources. This breakthrough removes traditional barriers to the adoption of AI and enables the widespread deployment of sophisticated reasoning systems in diverse organizational contexts.
This convergence of efficiency, performance, and accessibility represents a unique contribution that redefines the feasibility boundaries for the deployment of advanced AI systems on a scale.

\section{Discussion}

\subsection{Why Specialized Small Language Models Outperform Large General-Purpose Models}

Our findings reveal that task-specific optimization fundamentally outperforms scale-based approaches for tool-calling applications. The baseline models (ChatGPT-CoT, ToolLLaMA-DFS, ToolLLaMA-CoT, Claude-CoT) were trained on broad, general-purpose datasets that lack the specific tool-calling patterns and reasoning structures required for effective API manipulation. While these models excel at general language tasks, they struggle with the precise format requirements and multi-step reasoning chains essential for tool use.

The superior performance of our 350M parameter model stems from three critical factors: parameter efficiency, behavioral focus, and evaluation alignment. Large language models suffer from parameter dilution, where the vast majority of parameters are optimized for general language understanding rather than tool manipulation. Our SLM concentrates all its capacity on tool-calling behaviors, resulting in more efficient parameter utilization where billions of parameters in baseline models become a liability rather than an asset.

Furthermore, models trained on diverse datasets often exhibit overgeneralization, attempting to apply broad reasoning patterns that are suboptimal for structured tool calling. They generate verbose explanations or attempt creative solutions when precise, formatted API calls are required. Our SLM learned to suppress irrelevant behaviors and focus exclusively on the structured thought-action-observation patterns that lead to successful tool execution.

\subsection{Optimal Parameter-Task Alignment}

Our results demonstrate that 350M parameters represents a strategic sweet spot for tool-calling applications. This parameter count provides sufficient capacity to learn API interaction patterns, parameter mapping, and error handling without the complexity overhead that leads to inconsistent outputs in larger models. The capacity aligns precisely with the complexity requirements of tool-calling tasks, avoiding both underfitting (insufficient capacity) and overfitting (excessive complexity).

Tool calling requires structured reasoning patterns rather than creative language generation. Our targeted fine-tuning approach created a domain expert in tool calling, while baseline models remain generalists. This specialization enables more accurate decisions about API selection, parameter specification, and error handling—critical factors that directly impact pass rate performance.

\subsection{Limitations and Potential Concerns}

Despite the promising results, several limitations must be acknowledged:

\textbf{Generalization Beyond ToolBench:} Our model was specifically optimized for ToolBench evaluation criteria and may not generalize to other tool-calling frameworks or real-world API ecosystems with different interaction patterns. The tight coupling between training data and evaluation metrics raises questions about performance on novel tool domains.

\textbf{Limited Contextual Understanding:} The 350M parameter constraint, while optimal for tool calling, may limit the model's ability to understand complex contextual nuances or handle ambiguous user requests that require sophisticated reasoning before tool selection. Larger models may excel when tool calling is embedded within complex conversational contexts.

\textbf{Scalability to Complex Tool Ecosystems:} Our evaluation focused on a controlled set of APIs. Real-world applications often involve hundreds of interconnected tools with complex dependencies, authentication requirements, and error handling scenarios that may exceed our model's learned patterns.

\textbf{Training Data Dependency:} The model's performance is inherently limited by the quality and coverage of ToolBench training data. Biases, gaps, or outdated patterns in the training set directly impact the model's tool-calling capabilities, potentially creating brittle behavior when encountering novel API designs.

\textbf{Computational Resource Requirements:} While smaller than baseline models, the fine-tuning process still requires significant computational resources and high-quality training data, potentially limiting accessibility for organizations with limited ML infrastructure.

\textbf{Long-term Maintenance:} As APIs evolve and new tools emerge, the specialized nature of our model may require frequent retraining to maintain performance, whereas larger general-purpose models might adapt more readily to novel tool patterns through few-shot learning.

\subsection{Implications for Future Work}

These findings suggest that domain-specific optimization at moderate scale represents a viable alternative to the prevalent "scaling law" paradigm for specialized applications. Future research should investigate the generalization boundaries of specialized SLMs and develop hybrid approaches that combine the efficiency of targeted models with the adaptability of larger systems. The optimal parameter count likely varies across different specialized domains, warranting systematic investigation of task-complexity to model-capacity relationships.

\section{Conclusion}
This work demonstrates that Small Language Models, when trained with targeted strategies, can achieve agentic tool calling performance that significantly exceeds larger counterparts. Our fine-tuned OPT-350M model's 77.55\% pass rate represents a breakthrough in efficient AI deployment, proving that thoughtful design and domain-specific training can overcome traditional parameter-performance trade-offs.

The implications are significant for enterprise AI adoption, showing that organizations can deploy sophisticated AI capabilities without prohibitive infrastructure costs. Our results establish a pathway toward encouraging development of advanced AI capabilities by making them accessible, affordable, and deployable at scale.

Future work should explore the generalizability of our approach across different domains and investigate the theoretical foundations underlying the effectiveness of targeted fine-tuning for small models.

\end{document}